\documentclass{llncs}
\usepackage{algpseudocode}
\usepackage{algorithm}
\usepackage{subfig}
\usepackage{graphicx}
\usepackage{amsmath}
\usepackage{color}

\usepackage{verbatim} 

\begin{document}

\title{DASH: Dynamic Approach for Switching Heuristics}

\author{Giovanni Di Liberto\inst{1}, Serdar Kadioglu\inst{2},  Kevin Leo\inst{3}, Yuri Malitsky\inst{1}}
\institute{Cork Constraint Computation Centre, University College Cork, Ireland\\\email{dilibert@dei.unipd.it, y.malitsky@umail.ucc.ie} \and Oracle, United States \\\email{serdrk@gmail.com} \and Faculty of IT, Monash University, Australia \\\email{kevin.leo@monash.edu}}

\maketitle

\begin{abstract}

Complete tree search is a highly effective method for tackling MIP problems, and over the years, a plethora of branching
heuristics have been introduced to further refine the technique for varying problems. Recently, portfolio algorithms have
taken the process a step further, trying to predict the best heuristic for each instance at hand. However, the motivation 
behind algorithm selection can be taken further still, and used to dynamically choose the most appropriate algorithm for 
each encountered subproblem. In this paper we identify a feature space that captures both the evolution of the problem in
the branching tree and the similarity among subproblems of instances from the same MIP models. We show how to exploit 
these features to decide the best time to switch the branching heuristic and then show how such a system can be trained 
efficiently. Experiments on a highly heterogeneous collection of MIP instances show significant gains over the pure 
algorithm selection approach that for a given instance uses only a single heuristic throughout the search.

\end{abstract}


Mixed Integer Programming (MIP) is a powerful problem representation that is ubiquitous in the modern world. The 
problem is represented as the maximization of an objective function while maintaining the specified linear inequalities 
and restricting some variables to only take integer values while others are allowed to take on any real value.

\vspace{-0.5cm}

\begin{align*}
	Maximize:\: & c^{T}x \\
	\text{subject to :}\: & Ax \leq b \\
					& l \leq x \leq u \\
					& x_j \text{ integer } \forall j \in D \text{ , where } D \subseteq \{1..n\} 
\end{align*}

Through this straight forward formulation it is possible to define a wide variety of problems, ranging from 
scheduling~\cite{KaMaSaSaSe:11} to production planning~\cite{LaYv:06} to network design~\cite{BaMaMi:97} to auctions~\cite{ZuNi:01}
and many others.

In practice, these problems are typically approached using branch and bound or branch and cut techniques~\cite{Mi:70,PaRi:91}.
Here the main idea is to perform deterministic and inductive reasoning to lower the domains of the variables. When
this is no longer possible, a variable is selected and assigned a value based on some guiding heuristic. Once such a 
decision is made the search proceeds to function deterministically. If or when it is later found that the decision led to an 
infeasible or sub-optimal solution, the search backtracks, returning to the parent node to try an alternate assignment.

The key behind the success or failure of this complete search approach is the order in which the variables are selected
and the order the values are assigned to them. Choosing certain variables can significantly reduce the domains of all
other variables, allowing the deterministic analysis to quickly find a contradiction or determine that no improving solution 
can exist in the subproblem. Alternatively, choosing the wrong variables can lead to exponentially longer run times.

Due to the critical importance of the selection of the branching variable and value, there have been a number of heuristics
presented~\cite{AcKoMa:04,LiSa:97,Ac:07}. Several of them are based on simple rules, eg. Most/Least Infeasible Branching base their decisions on the variable's fractionality. Other heuristics, like Pseudocost Branching, can adapt over time while others, like Strong Branching, test which of the fractional candidates gives the best progress before actually committing to any of them. Finally, there are also hybrid techniques, eg. Reliability Branching, that put together the positive aspects of Strong Branching and Pseudocost Branching. A good overview of these and other heuristics can be found in~\cite{Ac:07}.


The efficiency of the search, however, can be much improved if we could use the correct heuristic at the correct time in
the search. Work with portfolios has already shown that there is often no single solver or approach that works optimally
on every instance~\cite{KaMaSeTi:10,MaHeHoNuSu:08,XuHuHoBr:12}. We also know that throughout the branching 
process, as certain variables get assigned and the domains of others are changed, the underlying structure of the
subproblems changes. In this paper, we show how to identify changes in the problem structure and therefore how to 
make a decision of when it is the best time to switch the employed guiding heuristic.

While a similar approach was recently introduced in~\cite{KaMaSe:12}, this work expands the research from the set
partitioning problem with problem dependent heuristics, to the much more general problem of MIP. We also provide a
detailed analysis of how the problem structure changes over time and clearly demonstrate the effectiveness of the 
proposed approach on real benchmarks that are of interest to the community.

\vspace{-0.3cm}
\section{Dynamic Switching}
\vspace{-0.1cm}

The objective motivating this work is to create a solver that dynamically adjusts its search strategy, selecting
the most appropriate heuristic for the subproblem at hand. In a high-level overview, we want the solver to perform a
standard branch and bound procedure, but before choosing the next branching variable and value, it will analyze the
structure of the current subproblem using a set of representative features. Using this structural information, the solver
would be able to predict that a specific heuristic is likely better than the alternatives, and employ it to make the next 
decision. We refer to such a strategy as a Dynamic Approach for Switching Heuristics (DASH).

The specifics of DASH are described in Algorithm~\ref{alg:DASH}. Modeled after the ISAC approach~\cite{KaMaSeTi:10}, 
DASH assumes that instances that have similar features share the same structure and so will yield to the same
algorithm. We will therefore employ clustering to identify these groups of instances. DASH is provided the current 
subproblem, the heuristic employed by the parent node, the centers of the known clusters, and the list of available 
heuristics. Because determining the feature can be computationally expensive and because switching heuristics at lower 
depths of the search tree has a smaller impact on the quality of the search, DASH only chooses to switch the
 guiding heuristic up to a certain depth and only at predetermined intervals, choosing the
parent's heuristic in all other cases. When a decision does need to be made, the approach computes the features of
the provided subproblem and determines the nearest cluster based on the Euclidean distance. In theory, any distance
metric can be used here, but in practice we found that Euclidean works well in the general case. In the end, DASH, 
employs the heuristic that has been determined best for that cluster.

\begin{algorithm}[t]
	\caption{DASH - branch callback}\label{alg:DASH}
	\begin{algorithmic}[1]
		\Procedure{branchCallback}{$subproblem,parent,centers,heuristics$}
			\If {$depth < maxDepth \;\;\textbf{and}\;\; depth \;\%\; interval==0$}
				\State ${x}\gets featuresComputation(subproblem)$
				\ForAll{center \textit{c} in \textit{cs}}
					\State ${distance}_{i}\gets euclideanDistance(x, centers)$
				\EndFor
				\State ${cluster}\gets argmin(distance)$ 
				\State ${heuristic}\gets heuristics_{cluster}$
			\Else
				\State ${heuristic}\gets parent.heuristic$
			\EndIf
				\State $ExecuteBranching(subproblem, heuristic)$
		\EndProcedure
	\end{algorithmic}
\end{algorithm}

As can be inferred from this algorithm, the key component that determines the success or failure of DASH, is the correct
assignment of heuristic to cluster. To train this, we follow a similar procedure first described in~\cite{KaMaSe:12}. For each instance
in the training set we compute an assortment of subproblems that are observed when using each of our heuristics.
This extended problem set allows us to get a better overview of the type of subproblems DASH will be encountering, 
as opposed to just using the original training instances. Computing the features of the extended problem set, we
cluster the instances. For this we employ g-means~\cite{HaEl:03}, a general clustering approach that automatically 
determines
the best number of clusters for the dataset in question. In particular, this clustering approach assumes that a good cluster
is one that has a gaussian distribution around the cluster center. Starting with all instances being in a single cluster,
g-means iteratively calls 2-means, to split a cluster in two. If the new clusters are more gaussian than the original,
the split is accepted and the procedure continues. Once all the instances are clustered, the clusters with fewer instances
than a certain threshold are absorbed by the nearest clusters.

Once all the subproblems are clustered, we have to determine which heuristic is best in which scenario. However, an
important caveat to this is that the decision of a using a heuristic for a certain cluster also affects all other decisions.
This is because DASH can switch heuristics several times, and the types of subproblems observed after applying one
heuristic will likely be different then when another one has been applied. Therefore, we employ the parameter tuner
GGA~\cite{AnSeTi:09} to simultaneously assign heuristics to all clusters, using only the original instances for training.

\vspace{-0.2cm}
\section{Experimental Setup}
\vspace{-0.1cm}

In order to set the stage for DASH, three things are necessary. First, we must have a descriptive feature set that can
correctly distinguish between different classes of instances, but also do this with minimal overhead. Second, there must
be a diverse set of heuristics each of which performs well on different kinds of instances. Finally, there must be a 
heterogeneous domain, with a large number of benchmark instances. We touch on all three of these components in
this section.

We implement our feature computation and heuristics through extending the state-of-the-art MIP solver 
Cplex version 12.5~\cite{Cplex}. Here, we only modify the built in branching strategy by implementing a branch callback 
function based on Algorithm \ref{alg:DASH}. Because all the tested approaches require this branch callback to be enabled,
the comparability of the results is guaranteed\footnote{Note that Cplex switches off certain heuristics as soon as branch 
callbacks, even empty ones, are being used so that the entire search behavior could be different.}. Finally, in order to
obtain reliable results, we run each Cplex execution in the single core version. The experiments were run on dual Intel
Xeon E5430 quad-core processors (2.66Ghz) computers with 12GB of DDR-2 FB-DIMM 667MHz memory.

\vspace{-0.1cm}
\subsection{Feature Space}

The features have to capture as many aspects of the problems as possible without becoming too expensive to compute.
To do this, we gather statistics about the problem definition of the remaining subproblem, a process similar to the one
employed in~\cite{KaMaSe:12}. Specifically, we compute:

\begin{itemize}
    \item Percentage of variables in the subproblem;
    \item Percentage of variables in the objective function of the subproblem;
    \item Percentage of equality and inequality constraints;
    \item Statistics (min, max, avg, std) of how many variables are in each constraint;
    \item Statistics of the number of constraints in which each variable is used;
    \item Depth in the branch and bound tree.
\end{itemize}

Wherever a feature has to do with the problem variables, we separately compute the same feature for each type of 
variable type: eg. continuous, integer, and binary. Therefore, the resulting set is composed of 40 features.

\subsection{Branching Heuristics}

In order to realize and test our solving approach, we implemented a portfolio of six branching heuristics.

\vspace{-0.2cm}

\subsubsection{Most Fractional Rounding (MF)} 
One of the simplest MIP branching techniques is to select the variable that has a relaxed LP solution whose fractional part
is most fractional and to round it first. The driving reasoning behind this is to make decisions on variables that deterministic
analysis is least certain about. Therefore, this heuristic strives to find infeasible solutions as quickly as possible.

\vspace{-0.2cm}

\subsubsection{Less Fractional Rounding (LF)} 
Alternatively to MF, this technique selects the the variable that has a relaxed LP solution whose fractional part is closest to
an integer value and to round it first. This is done to gently nudge the deterministic reasoning in whatever direction it is
currently pursuing, with a smallest chance of making a mistake.

\vspace{-0.2cm}

\subsubsection{Less Fractional And Highest Objective Rounding (LFHO)} 
This heuristic is based on the same motivation behind the Less Fractional Branching. For each subproblem we branch on
the variable for which the pair \emph{p=(fr, -obj)} is minimized (where \emph{fr} is the fractionality and \emph{obj} is the 
objective value). This means that, if we branch on a variable \emph{k} in [1,n], the following propriety is guaranteed: 

\vspace{-0.1cm}

\begin{equation*}
\forall i \in [1,n], \; fr_{k} < fr_{i} \: or \: obj_{k} > obj_{i}
\label{eq:heu_3}
\end{equation*}

\vspace{-0.2cm}

\subsubsection{Most Fractional And Highest Objective Rounding (MFHO)} 
We use a modification of the previous approach, but this time we focus on the most fractional variables. For each 
subproblem we branch on the variable for which the pair \emph{p=(fr, obj)} is maximized. In this case the guaranteed 
property is:

\vspace{-0.1cm}

\begin{equation*}
\forall i \in [1,n], \; fr_{k} > fr_{i} \: or \: obj_{k} > obj_{i}
\label{eq:heu_2}
\end{equation*}

\vspace{-0.2cm}

\subsubsection{Pseudocost Branching Weigthed Score (PW)} 
This heuristic is based on the pseudocosts, numerical values that estimates the variation in objective value for rounding up
or rounding down, called respectively up-pseudocost and down-pseudocost. The pseudocosts of a variable can be
combined in a score function (\ref{eq:heu2}) that returns a numeric value. This result is used to guide the branching, for 
which we choose the variable that maximize the score. Further details can be found in~\cite{AcKoMa:04}.

\vspace{-0.1cm}

\begin{equation*}
score(q^{-}, q^{+}) = (1-\mu)*min(q^{-}, q^{+}) + (\mu)*max(q^{-}, q^{+}), \;\;\mu = 1/6.
\label{eq:heu2}
\end{equation*}

\vspace{-0.2cm}

\subsubsection{Pseudocost Branching Product Score (P)} 
This approach is based on the same idea as PW. The difference lies in the score function that is now the product of the two pseudocosts.




\subsection{Dataset}

In order to obtain a solver that works well for a generic MIP problem we collected instances from many different datasets: 
\emph{miplib2010}~\cite{MIPLIB2010}, \emph{fc}~\cite{At:01}, \emph{lotSizing}~\cite{AtMu:04}, \emph{mik}~\cite{At:03}, \emph{nexp}~\cite{AtNeSa:01}, \emph{region}~\cite{LePeSh:00}, and \emph{pmedcapv}, \emph{airland}, \emph{genAssignment},
\emph{scp}, \emph{SSCFLP} were originally downloaded from~\cite{Saxena}. From an initial dataset of about 900 instances we filtered those for which all our solvers timed out in 1,800 seconds. We then removed the easy instances, solved entirely during the Cplex presolving or in less than 
one second by each solver. We finally obtained a dataset of 341 instances with the desired properties. We randomly 
selected 180 for the training set and 161 for the testing set. 

If we cluster our training data the distribution of instances per cluster can be seen in Table~\ref{table:featuresRoot}.
Each row is normalized to sum unto 100\%. Thus for Cluster~1, 25\% of the instances are from the \emph{airland} dataset.
From this table we first observe that there are not enough clusters to perfectly separate the different datasets into unique 
clusters. This, however, is not what we would want to see. This is because we are more interested in capturing similarities 
between instances, not splitting benchmarks. And we observe that the \emph{region100} and \emph{region200} instances 
are grouped together. We also see that Cluster 4 logically groups the \emph{LotSizing} and the \emph{SSCFLP} instances 
together. Finally, we see that the instances from the \emph{miplib}, those instances that are supposed to be an 
overview of all problem types, are spread across all clusters. 

This clustering therefore demonstrates that we both have a diverse set of instances and that our features are representative
enough to automatically notice interesting groupings.

\begin{table*}[t]
\setlength{\tabcolsep}{5pt}
\centerline{
\begin{tabular}{ccccccccccccc}
\hline
 & \textbf{1} & \textbf{2} & \textbf{3} & \textbf{4} & \textbf{5} & \textbf{6} & \textbf{7} & \textbf{8} & \textbf{9} & \textbf{10} & \textbf{11} & \textbf{12}\\
\hline
\textbf{Cluster} 1 & 20 & -     & -     & 25    & -     & 25    & -     & 30    & -     & -     & -     & -\\
\textbf{Cluster} 2 & -   & 45   & -     & -     & -     & 14    & 41    & -     & -     & -     & -     & -\\
\textbf{Cluster} 3 & -  & -     & -     & -     & 1     & 5     & -     & -     & 18    & 62    & 14    & -\\
\textbf{Cluster} 4 & -  & -     & 49    & -     & -     & 7     & -     & -     & -     & -     & -     & 44\\
\textbf{Cluster} 5 & -  & -     & -     & -     & 98    & 2     & -     & -     & -     & -     & -     & -\\
\hline
\end{tabular}
}
\caption{Instance distribution (percentage) in the clusterization at the root node. The problem types are:
1: airland, 2: fc, 3:GenAssignment, 4: LotSizing, 5: mik, 6: miplib2010, 7: nexp, 8: pmedcap, 9: region100, 10: region200, 11: scp, 12: SSCFLP}\label{table:featuresRoot}
\end{table*}

\setlength{\textfloatsep}{3pt}

\vspace{-0.3cm}
\section{Numerical Results}
\vspace{-0.1cm}

With the described methodology, the main question that needs to be addressed is whether switching heuristics can
indeed be beneficial to the performance of the solver. To test this, for each of the instances in our test set, we ran each
of the implemented heuristics without allowing any switching. We then also ran two versions of a solver that switched 
between heuristics uniformly at random. The first solver switched between all heuristics, while the second switched only 
among the top four best heuristics. The results are summarized in Table~\ref{table:prelim}.

What we observe is that neither of the random switching heuristics perform very well by themselves. However, based on
the performance of the virtual best solver\footnote{VBS is an oracle solver that for every instance always uses the strategy that results in the shortest runtime} that employs these new solvers, the performance can be further improved beyond
what is possible when always sticking to the same heuristic. The question therefore now, becomes, if we can get improved
performance just by switching between heuristics randomly, can we do even better if we do so intelligently?

\begin{table*}[t]
	\centering
	\setlength{\tabcolsep}{10pt}
	\begin{tabular}{l|cccc}
		\hline
 		\textbf{Solver} 		& \textbf{Avg} 	& \textbf{Par10}	& \textbf{\%Solved}\\	\hline
 		\textbf{BSS} 		& 315 		& 1321 		& 93.8	\\
 		\textbf{RAND 1} 	& 590 	   	& 4414 		& 77.0	\\
		\textbf{RAND 2} 	& 609 		& 5137 		& 72.7	\\
 		\textbf{VBS} 		& 225 		& 326 		& 99.4	\\
 		\textbf{VBS\_Rand} 	& 217 		& 217 		& 100	\\	\hline
	\end{tabular}
	\caption{Solving times on the testing set.}\label{table:prelim}
\end{table*}

\setlength{\floatsep}{1pt}

\begin{figure}[t]
	\centerline{
	\subfloat[Multi-path evolution]{\includegraphics[width=0.40\textwidth]{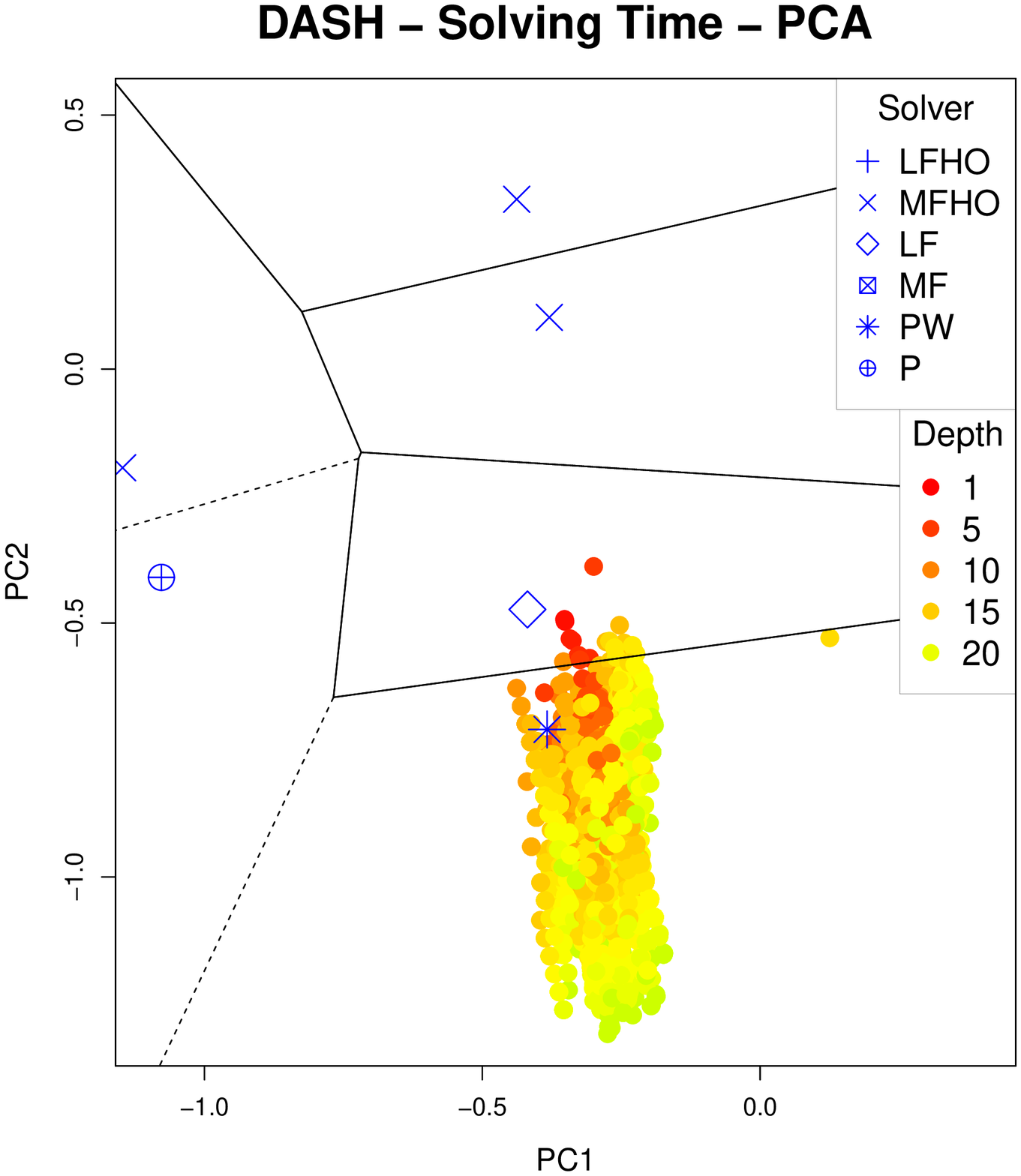}}
	\subfloat[Single-path evolution]{\includegraphics[width=0.40\textwidth]{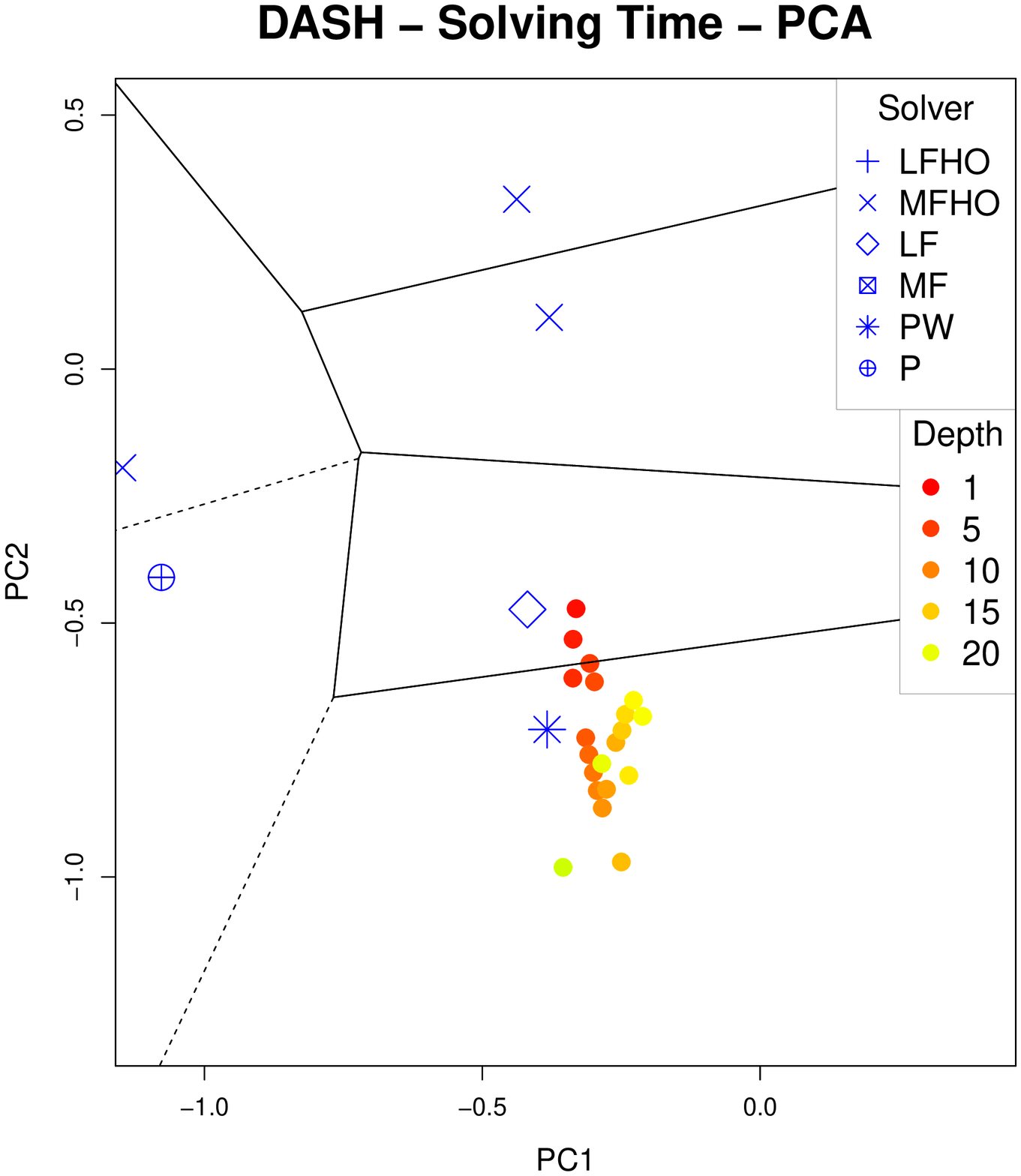}}
	}
	\caption{Position of a subproblem in the feature space based on depth.}
	\label{fig:depthInterval}
\end{figure}

To answer this question, we must first set a few parameters of our solver. Particularly, till what depth should we allow our
solver to switch heuristics, and at what interval? For this, we cluster the extended dataset that includes both the original
training instances and the possible observed subproblems. There are a total of 10 clusters formed. Projecting the feature
space into two dimensions using Principal Component Analysis (PCA)~\cite{AbWi:10} we present Figure~\ref{fig:depthInterval}. Here, the cluster boundaries are represented
by the solid lines, and the best heuristic for each cluster is represented by a unique symbol at its center. On these
figures, we also show the typical way in which features change as the problem is solved with a particular heuristic. The nodes are colored based on the depth of the tree, with (a) showing all the observed subproblems and (b) that of a single branch.

What this figure shows is that the features change gradually. This means
that there is no need checking the features at every decision node. We therefore choose to check the subproblem features
at every 3rd node. Similarly, the figure and those like it, show that using a depth of 10 is reasonable, as in most cases 
the nodes don't span across more than two clusters.

We use GGA to tune the parameters of DASH, computing the best heuristic for each cluster.  We then present the results in
Table~\ref{table:isac-dash} where we compare it to a vanilla ISAC approach that for a given instance chooses the single
best heuristic and then does not allow any switching. What we observe is that DASH is able to perform much better than
its more rigid counterpart. However, we do allow for the possibility that switching heuristics might not be the best strategy 
for every instance. We therefore also introduce DASH+, which first clusters the original instances using ISAC and then 
allows each cluster to independently decide if it wants to use dynamic heuristic switching.

\begin{table*}[t]
	\setlength{\tabcolsep}{10pt}
	\centering
	\begin{tabular}{l|cccc}
		\hline
 		\textbf{Solver} 	& \textbf{Avg} 	& \textbf{Par10}	& \textbf{\%Solved}\\	\hline
 		\textbf{BSS} 		& 315 		& 1321 		& 93.8	\\
 		\textbf{ISAC} 		& 302 		& 1107 		& 95.0	\\
        \textbf{ISAC\_filt}	& 289 		& 892 		& 96.3	\\
 		\textbf{DASH} 		& 251 		& 956 		& 95.7	\\
	    \textbf{DASH+} 	    & 255 		& 858 		& 96.3	\\
 		\textbf{DASH+filt} 	& \textbf{241} 		& \textbf{643} 		& \textbf{98.1}	\\
 		\textbf{VBS} 		& 225 		& 326 		& 99.4	\\
        \textbf{VBS\_DASH}	& 185 		& 286 		& 99.4	\\
        \hline
	\end{tabular}
	\caption{Solving times on the testing set.}\label{table:isac-dash}
\end{table*}

\setlength{\textfloatsep}{-5pt}


Taking a lesson from~\cite{KrMa:11}, which shows that often the features are not equally important, we tried to achieve better overall performance including a feature selection operation. In this paper we utilize the information gain filtering technique, often used in decision trees. In particular, this method is based on the calculation of entropy of the data as a whole and for each class. We apply the feature filtering to ISAC and DASH+ referring to them, respectively, as ISAC\_filt and DASH+filt, having an improvement in both cases. In particular, the resulting solver DASH+filt performs considerably better than everything else.

We finally show the performance of a virtual best solver if allowed to use DASH. And what we observe is that even though
the current implementation cannot overtake VBS, future refinements to the portfolio techniques will be able to achieve performances much better than techniques that rely purely on sticking to a single heuristic.

\vspace{-0.4cm}
\section{Conclusion}
\vspace{-0.3cm}

In this paper we introduce a Dynamic Approach for Switching Heuristics (DASH). Using MIP as the running example, we
show how to automatically determine when a subproblem observed during a branch and bound search is significantly 
different from what has been observed before, and therefore warrants a change of tactics used while solving it. Employing a diverse set of instances we demonstrate that significant performance improvements are possible if a solver does not stick to using a single guiding heuristic.


\bibliographystyle{splncs}
\bibliography{biblio.bib}

\end{document}